\documentclass[10pt,twocolumn,letterpaper]{article}

\usepackage[accepted]{cvpr}

\usepackage{multirow} 
\usepackage[utf8]{inputenc}

\definecolor{cvprblue}{rgb}{0.21,0.49,0.74}
\usepackage[pagebackref,breaklinks,colorlinks,allcolors=cvprblue]{hyperref}
\usepackage[utf8]{inputenc}

\title{RoleMotion: A Large-Scale Dataset towards Robust Scene-Specific Role-Playing Motion Synthesis with Fine-grained Descriptions}

\author{
Junran Peng*$^{1,3,5}$ \hspace{0.2cm} 
Yiheng Huang*$^{2,5}$ \hspace{0.2cm} 
Silei Shen*$^{1,5}$ \hspace{0.2cm} 
Zeji Wei$^{1,5}$ \hspace{0.2cm} 
Jingwei Yang$^{7}$ \hspace{0.2cm} 
Baojie Wang$^{1}$ \hspace{0.2cm} \\
Yonghao He$^{4}$ \hspace{0.2cm} 
Chuanchen Luo$^{5,6}$ \hspace{0.2cm} 
Man Zhang$^{2}$ \hspace{0.2cm} 
Xucheng Yin$^{1}$ \hspace{0.2cm} 
Wei Sui$^{\dag}$$^{4}$ \hspace{0.2cm} \vspace{0.2cm}\\
$^{1}$ University of Science and Technology Beijing \\ 
$^{2}$ Beijing University of Posts and Telecommunications \\
$^{3}$ Shunde Innovation School, University of Science and Technology Beijing \\
$^{4}$ D-Robotics 
$^{5}$ Linketic 
$^{6}$ Shandong University \\
$^{7}$ China University of Mining And Technology \\
}

\begin{document}

\twocolumn[{%
\renewcommand\twocolumn[1][]{#1}%
\maketitle
\includegraphics[width=\textwidth]{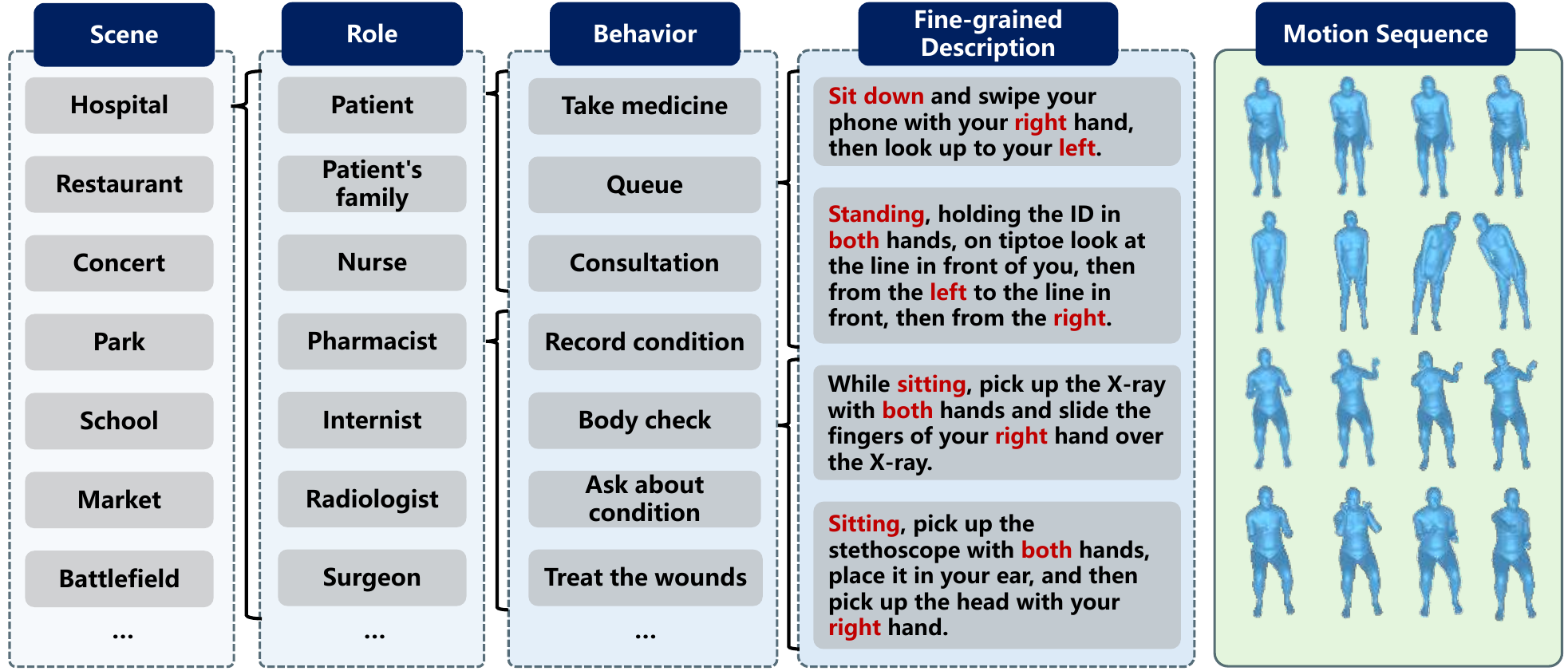}
\captionof{figure}{The design structure of {\bf RoleMotion}. {\bf RoleMotion} is a large scale scene-specific role-playing human motion datasets of high-quality compared to existing {\it text2motion} datasets. It aims to cover complete social activities in various scenes. All motions are carefully performed by actors and captured through MoCap devices instead of amassing existing datasets. The text annotations are fine-grained that body state, side of part and direction are required to be specified clearly.  \vspace{1em}}
\label{fig:teaser}
}]

\begin{abstract}


In this paper, we introduce {\bf RoleMotion}, a large-scale human motion dataset that encompasses a wealth of role-playing and functional motion data tailored to fit various specific scenes. 
Existing text datasets are mainly constructed decentrally as amalgamation of assorted subsets that their data are nonfunctional and isolated to work together to cover social activities in various scenes. 
Also, the quality of motion data is inconsistent, and textual annotation lacks fine-grained details in these datasets. 
In contrast, {\bf RoleMotion} is meticulously designed and collected with a particular focus on scenes and roles.
The dataset features 25 classic scenes, 110 functional roles, over 500 behaviors, and 10296 high-quality human motion sequences of body and hands, annotated with 27831 fine-grained text descriptions. 
We build an evaluator stronger than existing counterparts, prove its reliability, and evaluate various text-to-motion methods on our dataset. 
Finally, we explore the interplay of motion generation of body and hands. Experimental results demonstrate the high-quality and functionality of our dataset on text-driven whole-body generation. The dataset and related codes will be released.
\end{abstract}    
\section{Introduction}
\label{sec:intro}

\begin{figure}[!htp]
    \centering
    \includegraphics[width=0.47\textwidth]{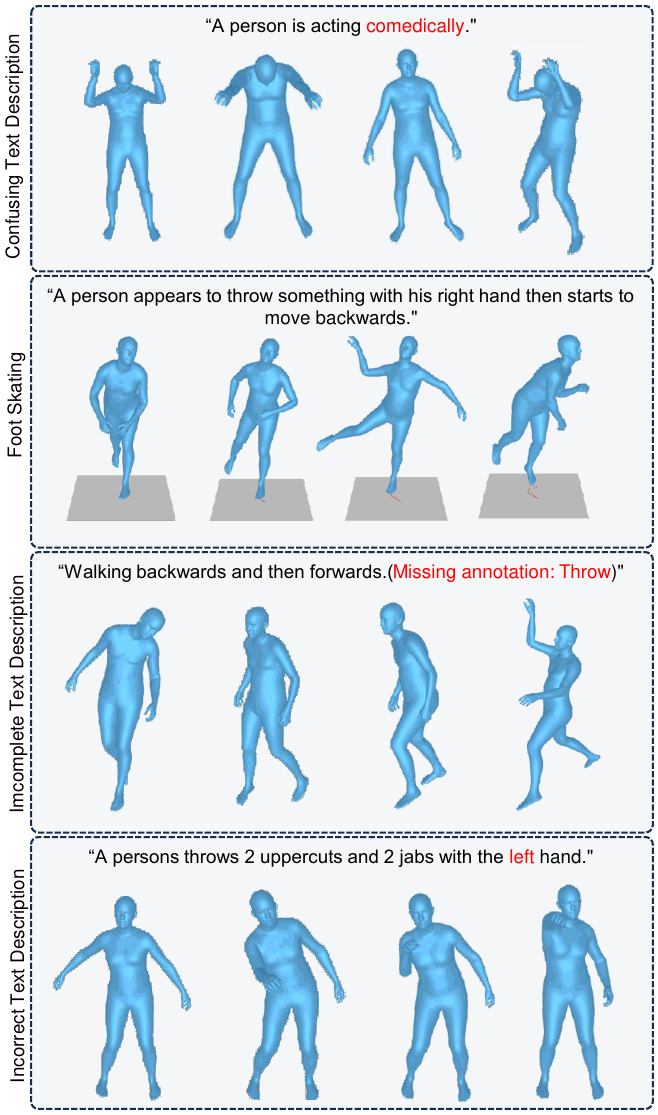}
    \vspace{-1mm}
    \caption{The data quality of HumanML3D raises significant concerns. More cases could be found in the supplementary material.}
    \vspace{-1mm}
    \label{fig:bad_cases_in_humanml3d}
\end{figure}

Creating immersive scenes with NPCs that exhibit natural human-like social behavior has long been a goal in virtual environment research. Human motion synthesis—generating natural, diverse motions from various conditional inputs—is essential to realizing this vision. Existing approaches generate gestures from speech~\cite{liu2022investigating,mao2019learning}, motions from action categories~\cite{cervantes2022implicit,guo2020action2motion,lucas2022posegpt,petrovich2021action}, dance from music~\cite{gong2023tm2d,siyao2022bailando,tseng2023edge,zhou2023ude}, and daily activities from text~\cite{chen2023executing,guo2022Text2Motion,guo2022tm2t,petrovich2022temos,zhang2022motiondiffuse}. However, progress in motion generation lags far behind that in image and video synthesis. A key bottleneck is the scarcity of high-quality motion data, which serves as an intermediate representation, unlike images and videos as final media forms. While monocular motion capture methods like~\cite{lin2023motionx} can extract pseudo-labels from online videos, their quality remains unreliable. Current text-to-motion datasets such as HumanML3D~\cite{guo2022Text2Motion} largely derive from amalgamations of earlier MoCap datasets~\cite{mahmood2019amass}. Similarly, HumanAct12~\cite{guo2020action2motion} builds on NTU-RGB-D~\cite{shahroudy2016ntu}, and AMASS combines 15 small motion datasets. Merging datasets with heterogeneous standards inevitably affects quality: as Fig.~\ref{fig:bad_cases_in_humanml3d} shows, HumanML3D contains noticeable foot-skating and annotation errors. Moreover, the unstructured and non-purposive distribution of these datasets hinders the generation of coherent motions that depict integrated scenario events.

To tackle the aforementioned problems, we propose {\bf RoleMotion}, a large-scale human motion dataset that contains scene-specific role-playing 3D motion sequences with fine-grained textual descriptions. 
We take scenes and roles into account when designing the dataset structure.
Firstly, we collect commonly used scenes,  such as {\it hospital}, {\it school}, {\it restaurant}, {\it market}, and enumerate all the essential roles within them.
Subsequently, we thoroughly list possible behavior types of each role in the scene and expand these behavior types to numerous specific motion sequences with fine-grained descriptions. 
In total, there are 25 classic and common scenes, 110 roles, more than 500 meta behaviors, 10296 specific motion sequences, and 27831 text descriptions in {\bf RoleMotion}. 
The data consists of both motion of body and hands while most existing {\it text-to-motion} datasets such as HumanML3D~\cite{guo2022Text2Motion} and KIT-ML~\cite{Plappert2016} only contain data of body motion. 
Besides, our text description is annotated precisely, including the body states, part-level actions, action directions, and even amplitude-related adjectives. 

To demonstrate the effectiveness of our motion data and fine-grained textual annotation, we benchmark existing open-sourced {\it text-to-motion} generation methods on our datasets. 
We preferentially select methods that are stable and popular, including
Motion Diffusion Model(MDM)~\cite{shafir2023human}, Motion Latent Diffusion(MLD)~\cite{chen2023executing}, MotionDiffuse~\cite{zhang2022motiondiffuse} and StableMoFusion~\cite{huang2024stablemofusion} as representatives of diffusion-based modelss, T2M-GPT~\cite{zhang2023generating}, MoMask~\cite{guo2024momask} and Text-to-Motion Retrieval(TMR)~\cite{petrovich23tmr} as representatives of auto-regressive methods. Experiments demonstrate the validity of our data, that various methods could be able to generate high-quality motion sequences well aligned with fine-grained textual descriptions.
To validate the quality, expressiveness and effectiveness of 
human motion generators as well, a reliable evaluator is crucially needed. However, the existing evaluator that is most commonly used~\cite{guo2020action2motion} is frequently challenged to be unconvincing and out-dated that the calculated scores could not reflect the real quality of generated motions for two possible reasons. In this work, we train a strong evaluator on our dataset and verify that it is much more convincing than its counterpart. All the metric scores in this work are calculated by our evaluator. 

\begin{table*}[h]
  \caption{Comparisons between RoleMotion and existing text-motion datasets. {\bf B}, {\bf H}, {\bf F} denote {\bf B}ody, {\bf H}and and {\bf F}ace. {\bf A} and {\bf C} denote that the dataset is an {\bf A}ggregation of existing datasets or {\bf C}ollected from scratch. {\bf GT} means that the motion annotations come from ground-truth of motion capture, while pseudo labels are predicted by models.}
  \centering
  \tabcolsep=15pt
  \resizebox{2.05\columnwidth}{!}{%
  \begin{tabular}{l|lll|ll}
    \toprule
    Dataset & Clips &  Text Annotations &  Motion-Annotations & Source & Annotation Type\\
    \toprule
     KIT-ML~\cite{Plappert2016} & 3911 & 6278 & B & A & GT\\
     AMASS~\cite{mahmood2019amass} & 11265 & - & B & A & GT\\
     BAREL~\cite{punnakkal2021babel} & 13220 & 91408 & B & A & GT\\
     HumanML3D~\cite{guo2022Text2Motion} & 14616 & 44970 & B & A & GT\\
     MotionX~\cite{lin2023motionx} & 81084 & 81084 & B, H, F & A  & Pseudo label\\
    \toprule
     {\bf RoleMotion(Ours)} & 10296 & 27831 & B, H & C & GT\\
    \toprule
  \end{tabular}
}
\vspace{-2mm}
\label{tab:compare}
\end{table*}

Whole-body motion generation involves the synthesis of motions of the body and hands.
Body movement generally dominates motion semantics, while hands are more dexterous and almost idle in many cases.
Since motion data is expressed in format of rotation instead of distance, the fact that hands own more joints than body but contribute less to visual representation raises a concern: Is it optimal to synthesize the motion of the body and hands by a single generator?
In this work, we explore the interplay of motion synthesis of the body and hand parts and suggest that it is better to train and evaluate the motion generation of the body and hands separately.


To summarize, our contributions are as follows:
\begin{itemize}
    \item We propose a large-scale human motion dataset that contains high-quality scene-specific role-playing motion data of body and hands, with fine-grained text descriptions.  
    \item We build a strong evaluator more reliable than existing counterparts, prove its validity and build a new {\it text2motion} benchmark. 
    \item We explore the interplay of motion generation of body and hands, reveal insights in terms of training and evaluation of whole body motion synthesis. 
\end{itemize}

\section{Related Work}
\label{sec:related-work}


\subsection{Human Motion Datasets}
Human motion modeling is a long-standing problem in computer graphics.
Early datasets~\cite{accad_motion_lab,troje2002decomposing,de2009guide,mandery2015kit,mocapdata_website,sigal2010humaneva,muller2007documentation,akhter2015pose,loper2014mosh,sfu_mocap_database,hoyet2012sleight,trumble2017total} aims at action recognition and motion capture, with annotations of 3D joints collected through optical marker or IMU.
These datasets apply a variety of skeleton standards that the number and location of markers for motion capture are different, and AMASS~\cite{mahmood2019amass} unifies them by mapping into the common SMPL format.
KIT Motion-Language(KIT-ML)~\cite{Plappert2016} is the first motion-language datasets that each motion data is annotated with text description, enabling the task of text to motion generation. 
HumanML3D~\cite{guo2022Text2Motion} combines the dataset of HumanAct12~\cite{guo2020action2motion} and the annotated AMASS, becoming the most popular {\it text2motion} dataset. MotionX~\cite{lin2024motion}, a recently proposed dataset, collects data from online video and annotates 3D joints based on the pseudo label generated by monocular motion capture method. These datasets have greatly advanced the development of human motion generation, spawning plenty of remarkable methods. 

However, since most existing datasets are amalgamations of assorted motion capture datasets, the overall data quality is far from satisfactory. The cases of footskate and distorted motions frequently occur, and text annotations are coarse-grained, ambiguous, and sometimes incorrect, especially when indicating which body parts are oriented in which specific directions. More importantly, data in these datasets is collected without a unified purpose or application motivation, making it difficult for these motions to adequately represent a specific scene or play. To address the issues, we develop a large-scale expressive human motion dataset from the perspective of covering all the role-playing in various scenes.

\subsection{3D Human Motion Synthesis}
Thanks to these human motion datasets, tremendous progress has been witnessed recently in the field of 3D human motion generation. There are methods that could generate motion sequences based on action categories~\cite{liu2022investigating,mao2019learning}, audio~\cite{gong2023tm2d,siyao2022bailando,tseng2023edge,zhou2023ude} and text prompts~\cite{chen2023executing,guo2022Text2Motion,guo2022tm2t,petrovich2022temos,zhang2022motiondiffuse}.
~\cite{guo2022Text2Motion} proposes to learn an aligned embedding space between natural language and human motions to enhance RNN model.
In ~\cite{wang2020learning,cai2018deep,guo2022action2video,petrovich2021action}, GAN is applied to progressively expand the partial motion sequence to a complete action.
Similarly, MotionCLIP~\cite{tevet2022motionclip} aligns text and motion embedding using CLIP~\cite{tevet2022motionclip} as the text encoder and rendered images as extra supervision.
T2M~\cite{guo2022Text2Motion} resorts to VAE to map the text prompt into a normal distribution of human motion. 
T2M-GPT~\cite{zhang2023generating} propose a 2-stage framework that combines VQVAE and GPT. 
Following the trends in area of image and video generation, diffusion-based models become popular.
MDM~\cite{tevet2023human} and MotionDiffuse~\cite{zhang2022motiondiffuse} both introduce versatile and controllable motion generation frameworks that could generate diverse
 motions with comprehensive texts.
MLD~\cite{chen2023executing} utilizes latent embeddings for efficient motion generation.
In StableMoFusion~\cite{huang2024stablemofusion}, model architecture, training recipes, and sampling strategies are exhaustively analyzed, and a stable and efficient motion generator is introduced.
MoMask~\cite{guo2024momask} introduces a masked modeling framework to fill up missing tokens from empty sequence. 
In this paper, we train a stronger evaluator compared to the commonly used one~\cite{guo2020action2motion}, validate the reliability of our evaluator, and benchmark popular open-sourced human motion generation methods mentioned above on our dataset.

\subsection{Agent-Based Role-Playing}
Creating believable agents and intellectual NPCs to craft immersive interactive experiences in animations or sophisticated game has always been a facinating dream for human. 
RoleLLM~\cite{wang2023rolellm} exhibits strong ability of agent to imitate talking style of any role, requiring only role descriptions and catchphrases for effective adaptation.
In ~\cite{park2023generative}, multiple agents of distinctive background and personality live and interact vividly in a small town driven by LLM. ~\cite{qian2023communicative} introduces a framework for agents to collaborate to run a coding company. 
Nevertheless, these agent-based activities are only conducted conceptually and exist merely on textual level. Precisely for this reason, our dataset is proposed to concretize them into visual representations. 
\section{RoleMotion}
\label{sec:dataset}

\begin{figure*}[t]
    \centering
    \includegraphics[width=1.0\textwidth]{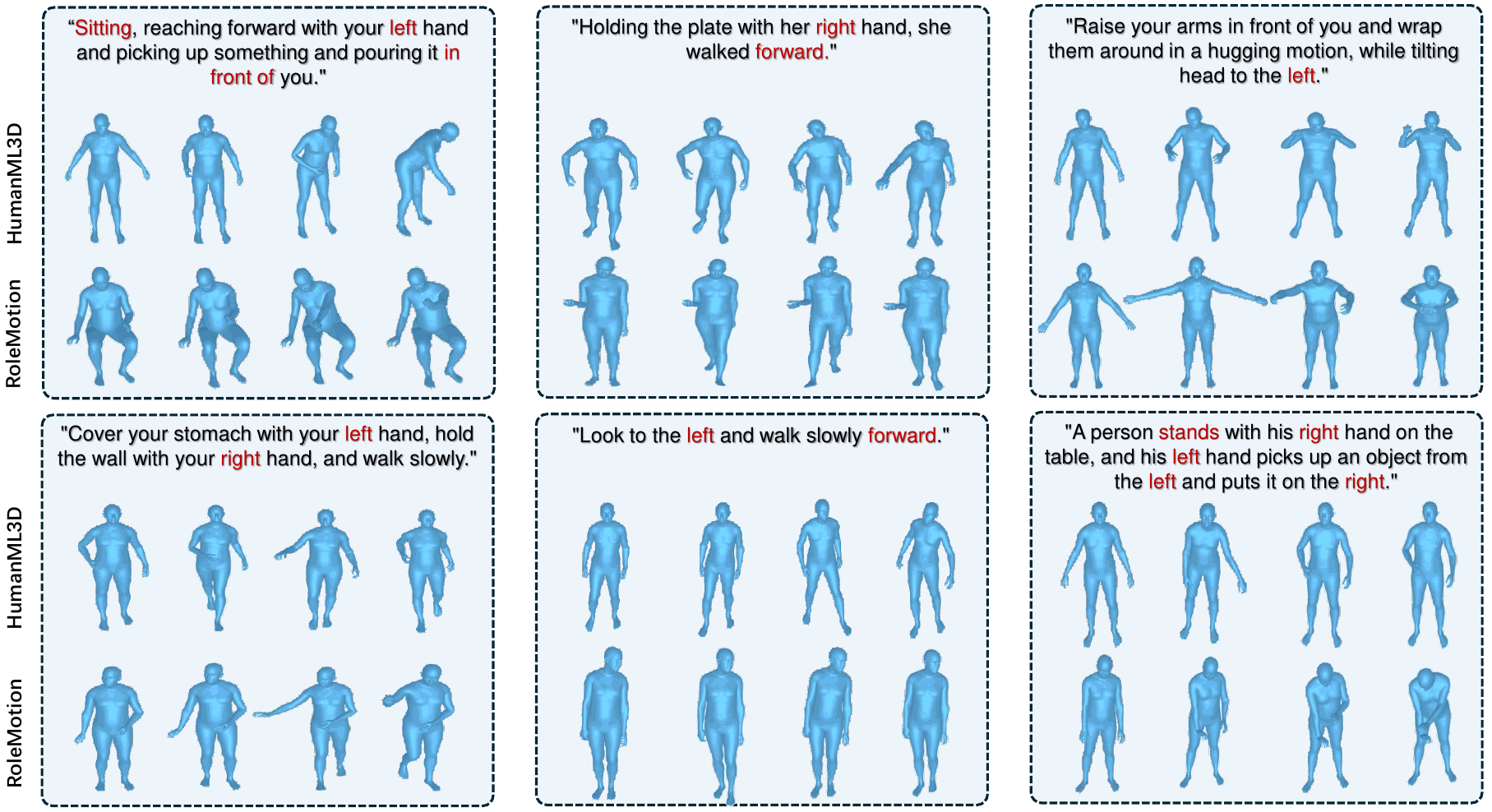}
    \vspace{-3mm}
    \caption{Qualitative results of models trained on HumanML3D and RoleMotion. Compared to HumanML3D, the motion data in RoleMotion is of high-quality and textual annotations are fine-grained that body state, side of part, and direction are required to be stipulated. 
    Almost all the action details specified in the text are accurately completed by model trained on RoleMotion, except the case in the top right that both models fail to correctly perform ``\textcolor{blue}{\it tilting head to the left}". Motions of model trained on HumanML3D tend to be unnatural and contain many errors at the detail level.  
    All the action categories in figures are included in both datasets. }
    \vspace{-3mm}
    \label{fig:fine_grained}
\end{figure*}

\subsection{Overview and Data Design}
Since datasets like HumanML3D and KIT-ML originate from existing motion capture datasets, it is unable to determine the composition of data. Oppositely, RoleMotion is designed towards specific purpose of covering complete social activities across various scenes from the beginning. As shown in Tab. ~\ref{tab:compare}, the 10296 motion sequences in RoleMotion are collected from scratch through MoCap devices instead of amassing existing datasets or predicting pseudo labels from existing videos, without any concerns about license.   
In addition, there is no need to unify different skeleton standards~\cite{mahmood2019amass} or obtain pseudo label from prediction~\cite{lin2023motionx}, which inevitably harms the quality of motion data.  

We take great consideration on scenes and roles when designing the structure of our dataset.
Firstly, we list scenes that are commonly seen in animations, video games, meta-verse, etc. 
Then we identify essential roles in these scenes and enumerate possible behaviors of each role. Ultimately, behaviors of each role are expanded to a variety of specific motion description. 
For instance, ``\textcolor{blue}{\it A singer singing in vocal concert}" is expanded to 88 specific motion entries like ``\textcolor{blue}{\it Holding the microphone in the left hand close to the mouth, while slightly swaying the body from side to side and singing}" or ``\textcolor{blue}{\it Holding a microphone in one hand and spreading the other hand outward while performing a song}". 
Noticeably, all the text descriptions are fine-grained, that body states like sitting and standing, the side of part and the moving direction are required to be explicitly specified. 
Totally, there are 25 classic scenes, 110 functional roles, more than 500 behaviors, 10296 high-quality human motion sequences including body and hands, and 27831 fine-grained text descriptions in {\bf RoleMotion}.


\subsection{Dataset Collection}
During data collection, 6 actors are employed to perform motions strictly following the fine-grained descriptions. Motion clips that do not strictly conform to the textual descriptions or are performed unnaturally are removed and re-captured on the spot. 
Xsens motion capture suit and Manus gloves are adopted for body and hand capturing, respectively. The body data captured using Xsens suits is in the format of 
Filmbox (FBX)
human skeletons of 23 joints, and hand skeleton has 30 joints. 

The occurrence of footskate that foot move around while contacting the ground may happen in data captured through any devices, thus we manually refine the data with footskate to ensure the high-quality of the dataset.
To contribute to both academic and industrial communities, we retarget all the motion sequences from Xsens standards(23+30 joints) to SMPL-H~\cite{pavlakos2019expressive}(22+30 joints) and Manny(24+30 joints) in Unreal5~\cite{unreal_engine_website}.
In addition, each text description is augmented to multiple pieces via ChatGPT~\cite{an2023chatgpt} to enhance the generalizability of data. Annotators would manually review the generated text and remove the pieces that could not ensure the integrity and accuracy of the original information.

\subsection{Evaluation Metrics}
\label{subsec:evaluator}
Previous methods are evaluated following the setting proposed in Text2Motion~\cite{guo2022Text2Motion}, that a motion feature extractor and text feature extractor are trained under contrastive loss to align the two spaces. Frechet Inception Distance(FID) score is calculated between generated motions and the ground-truth motions to indicate the quality. R-Precision is calculated through a model trained to align text and motion, which demonstrates the matching rate between text and motion. {\it Diversity} and {\it Multimodality} are also applied following the metrics in ~\cite{guo2020action2motion}. 
However, this evaluator is challenged to be weak and out-dated recently that these scores could not reflect the real quality of generated motions for two possible reasons. 
The model is based on a small RNN network of limited capacity, and it is trained on HumanML3D dataset that the data diversity and quality are far from satisfactory.
We follow the evaluation schema in TEMOS~\cite{petrovich2022temos} and TMR~\cite{petrovich23tmr}, training a transformer-based motion and text encoders on our dataset of fine-grained textual annotation and high-quality motion sequences. 
The decoder part in TEMOS is abandoned in our implementation, because we find the supervision of Kullback-Leibler(KL) divergence loss reducing the disparity of data which greatly hinders the validity of FID. 

Replacing the most commonly used and widespread evaluation methods needs to be down with caution.
One convincing approach to prove the effectiveness of our proposed method is to compare the R-Precision of ground-truth data among evaluators.
We compare performance of the original RNN-based evaluator~\cite{guo2022Text2Motion} trained on RoleMotion, and our transformer-based evaluator trained on RoleMotion. Tab.~\ref{tab:evaluator} shows that our evaluator outperforms the others obviously, and is utilized for all our experiments in this paper. 
\section{Motion Generation Benchmark}

\subsection{Baseline Methods}
\label{subsec: baseline}

\noindent\textbf{StableMoFusion.} utilizes Conv1D UNet with linear cross-attention for diffusion-based generation. For our dataset, we specifically adjusted the sampling steps of the DPMSolver~
\cite{lu2022dpm} to optimize performance.

\noindent\textbf{MotionDiffuse.} utilizes cross-modality linear transformer for diffusion-based generation. Instead of predicting noise, we adapt it to directly predict real samples.

\noindent\textbf{Motion Diffusion Model (MDM).} utilizes Transformer encoder for diffusion-based generation. For our dataset, we incorporate a \textit{padding\_mask} to make the Transformer independent of motion padding.

\noindent\textbf{Motion Latent Diffusion (MLD).} utilizes VAE to obtain latent motion encodings, followed by a transformer with long skip connections for diffusion in the latent space. During VAE training, we use the joint positions inferred from SMPL-H to compute the joint reconstruction loss.

\noindent\textbf{Text Motion Retrieval (TMR).} utilizes a Transformer-based joint synthesis and retrieval framework. In line with this method, we apply DistilBERT~\cite{sanh2019distilbert} and MPNet~\cite{song2020mpnet} to our text data, extracting token and sentence embeddings.

\noindent\textbf{MoMask and T2M-GPT.} both employ discrete representations to process continuous human motion data. During VQVAE training, we use the representation difference between adjacent frames as velocity to calculate the velocity reconstruction loss, and adjust the window size according to our dataset. When training on our full-body data, the codebook size is set to double that used for body-only data.

\subsection{Body\&Hand Motion Generation}
Whole-body motion generation involves the synthesis of temporal movement of the body and hands.
The body movement generally dominates the motion semantics described by text instructions in our setting.
In comparison, the hands are more dexterous and keep almost idle in many cases.
Their movement should be coordinated with body movement and mainly make effects in interaction scenarios.
Due to such a discrepancy, it is suboptimal to synthesize the motion of the body and hands by a single generator.
In light of this observation, we propose to disentangle the generation of body and hand motion.

\begin{figure}[!htp]
    \centering
    \includegraphics[width=0.47\textwidth]{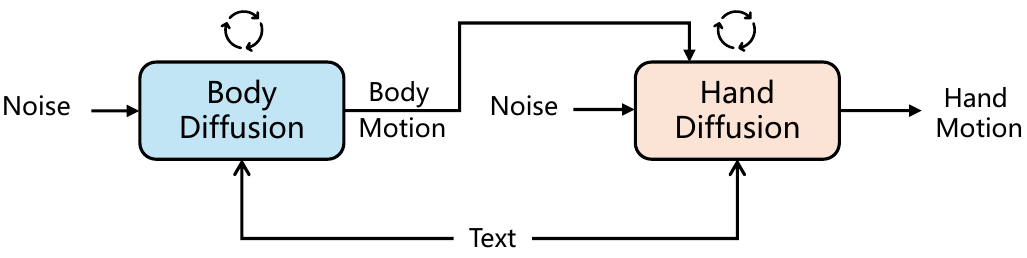}
    \vspace{-1mm}
    \caption{A pipeline of two-stage body\&hands motion generation. }
    \vspace{-1mm}
    \label{fig:two-stage}
\end{figure}

We employ efficient StableMofusion as the baseline and adapt it to two-stage generation.
The first stage focuses on generating body motion according to the text instruction, which is the same as the original setting of StableMofusion. 
We therefore directly invoke StableMofusion and train it on our dataset variant that excludes hand motions.
The second stage concentrates on the generation of hand motion.
Considering the dependency between hands and body, we condition the denoising network on the body motion.
We concatenate the ground-truth body motion with noisy hand motion as the input of the network during training.
Such ideal conditioning suffers from somewhat inconsistency between training and inference, which may lead to inferior synthesis quality of hand motion. 
To enhance the robustness of generation, we follow the practice in cascade diffusion~\cite{ho2022cascaded} and manually distort the body motion by Gaussian noise before conditioning.   
During inference, we invoke the first stage to generate the body motion. 
The outcome acts as the condition of the second stage for the synthesis of hand motion.
The outputs of both stages are combined to form the final results.

\section{Experiments}

\subsection{Implementation Details}

\paragraph{Dataset Split.}
We follow HumanML3D~\cite{guo2022Text2Motion} and MotionX~\cite{lin2024motion} to split the dataset into training, validation, and test sets with an 0.80 : 0.15 : 0.05 ratio. This division resulted in 8,236 samples for the training set, 515 samples for the validation set, and 1,545 samples for the test set.

\vspace{-2mm}
\paragraph{Motion Representation.}
In our work, we use the translation of the root joint and the rotation of all joints to represent the human body pose. Specifically, the i-th pose in the human motion sequence $m_i$ is defined by a tuple $(r^x,r^y,r^z,j^r)$, where $(r^x,r^z) \in \mathbb{R}$ are root translation on horizontal XZ-plane; $r^y \in \mathbb{R}$ is root height;
 $j^r \in \mathbb{R}^{6N}$ are $6D$ continuous rotations of all joints relative to their parent in a hierarchical structure, $N$ denoting the number of joints. Our data follows the skeleton structure of SMPL-H~\cite{pavlakos2019expressive} with 22 joints for the body and 30 joints for the hand. 
 
\paragraph{Data Pre-processing.}
We adjust the root xz-translations $(r^x,r^z)$ for each pose to be relative to the 0-th pose and reset the root y-translation $r^y$ to place the motion on the ground by aligning the motion sequence's lowest joint y-translation to zero based on SMPL-H kinematics. All motions are scaled to 20 FPS.

\paragraph{Training and Inference Details.}
For training, we normalize the motion data by calculating the mean and variance, then average the variances within translation and rotation dimensions separately to achieve two unified scales: one unit variance for translation, and another for rotation. In our experiments, we largely follow the original method's training and inference settings. All experiments are conducted on NVIDIA A40 GPUs.
More implementation details are provided in the appendix file.

\begin{table*}[t!]
 \caption{R-Precision of ground-truth from RoleMotion tested by different evaluators. B and H denote {\bf B}ody and {\bf H}and. The higher matching score of texts and motions is achieved by our evaluator indicating that our evaluator is more reliable.} 
 \vspace{-3mm}
\label{tab:evaluator}
  \centering
\resizebox{0.80\linewidth}{!}{
  \begin{tabular}{lcccccc}
    \toprule
    \multirow{2}{*}{\centering Evaluator} & 
    \multicolumn{3}{c}{R-Precision(Batchsize=32)$\uparrow$ }& 
    \multicolumn{3}{c}{R-Precision(Batchsize=256)$\uparrow$ }
    \\
    \cline{2-7}
    & top1 & top2 & top3 & top1 & top2 & top3 \\
    \midrule
     Text2Motion(B) & 
    $ 0.820^{\pm{ .003}}$ &$ 0.931^{\pm{ .001}}$ &$ 0.964^{\pm{ .001}}$ &		
     $ 0.509^{\pm{ .005}}$ &$ 0.680^{\pm{ .004}}$ &$ 0.768^{\pm{ .005}}$ \\
     Ours(B) & 
    $ \textbf{0.939}^{\pm{ .001}}$ &$\textbf{0.978}^{\pm{ .001}}$ &$ \textbf{0.987}^{\pm{ .001}}$ &		
     $ \textbf{0.802}^{\pm{ .005}}$ &$ \textbf{0.902}^{\pm{ .002}}$ &$ \textbf{0.931}^{\pm{ .001}}$ \\
     \midrule
     Text2Motion(B\&H) & 
     $ 0.867^{\pm{ .004}}$ &$ 0.950^{\pm{ .002}}$ &$ 0.974^{\pm{ .001}}$&		
    $ 0.608^{\pm{ .005}}$ &$ 0.770^{\pm{ .002}}$ &$ 0.838^{\pm{ .004}}$ \\
     Ours(B\&H) & 
     $ \textbf{0.954}^{\pm{ .002}}$ &$\textbf{0.988}^{\pm{ .001}}$ &$ \textbf{0.995}^{\pm{ .001}}$ &
     $\textbf{0.828}^{\pm{ .004}}$ &$ \textbf{0.926}^{\pm{ .002}}$ &$ \textbf{0.950}^{\pm{ .001}}$ \\
    \bottomrule
    \vspace{-5mm}
  \end{tabular}

}
 
\end{table*}

\begin{table*}[th]
	\caption{Higher R-Precisions are achieved on RoleMotion dataset by different evaluators indicating that the matching quality of text and motion in RoleMotion is much higher than HumanML3D.}
	\vspace{-0.3cm}
	\centering 
	\footnotesize
    \resizebox{0.85\linewidth}{!}{
	\begin{tabular}{llcccc}
    \hline
    Evaluator & Dataset  & R-Precision(Top1) & R-Precision(Top2) & R-Precision(Top3) & Diversity  \\
    \hline
    \hline
    Ours & HumanML3D  & 0.505 &0.645 & 0.711 & 15.050   \\
    \hline
    Ours & RoleMotion &  {\bf 0.939} &  {\bf 0.978} & {\bf 0.987} & {\bf 19.121}\\
    \hline
    Text2Motion & HumanML3D    & 0.511  & 0.703 & 0.797 & 9.503   \\
    \hline
    Text2Motion & RoleMotion   &  {\bf 0.820}  & {\bf 0.931} & {\bf 0.964} & {\bf 14.605}\\
    \hline
	\end{tabular}
    }
	\label{tab:evaluator_on_different_data}
\end{table*}

\subsection{Evaluators}
To validate the quality, expressiveness and effectiveness of human motion generators and datasets as well, it is essential to provide a reliable evaluator. 
As introduced in Sec.~\ref{subsec:evaluator}, we compare our evaluator with the most widely used one proposed in Text2Motion~\cite{guo2022Text2Motion}, based on the R-Precision of ground-truth.
Similar to TMR~\cite{petrovich23tmr} and HumanTomato~\cite{lu2023humantomato}, we train a text encoder and a motion encoder for evaluation but discard the motion decoder, and the comparative loss of InfoNCE~\cite{oord2018representation} with negative samples is utilized to better construct the latent space. Token embeddings and sentence embeddings are extracted using {\it distilbert-base-uncased} and {\it all-mpnet-base-v2} from our text annotations. 
To evaluate motion of body and body with hands respectively, we train evaluators on data of both settings.
As shown in Tab.~\ref{tab:evaluator}, our evaluators outperforms the counterpart in every settings. Given this, all the FID and R-Precision in this paper are calculated by our evaluators instead.
As illustrated in Tab.~\ref{tab:evaluator_on_different_data}, the RoleMotion dataset demonstrates significantly superior text-motion alignment compared to the HumanML3D dataset.

\subsection{Human Motion Synthesis}

\begin{figure}[!htp]
    \centering
    \vspace{-2mm}
    \includegraphics[width=0.47\textwidth]{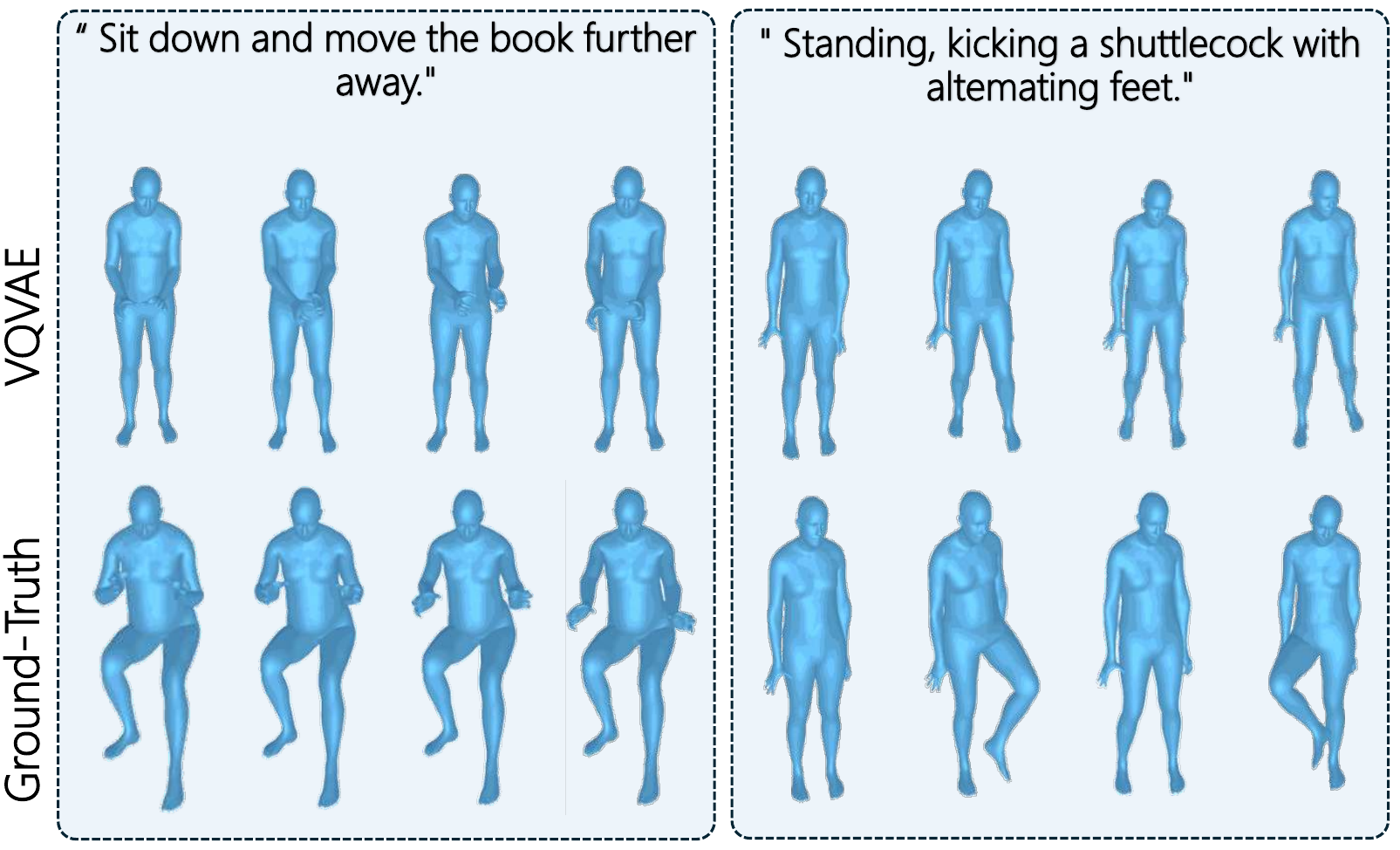}
    \vspace{-2mm}
    \caption{Recon. results of VQVAE trained on RoleMotion. }
    \vspace{-2mm}
    \label{fig:vqvae}
\end{figure}

Here we focus on two objectives: i) building a benchmark that shows how different methods perform on RoleMotion; ii) exploring the influence of generating hand motion while generating body motion.
We train 4 diffusion-based methods~\cite{tevet2023human, chen2023executing,zhang2022motiondiffuse,huang2024stablemofusion}, 
VAE-based TMR~\cite{petrovich23tmr}, VQVAE-based methods MoMask~\cite{guo2024momask} and T2M-GPT~\cite{jiang2023motiongpt} on our datasets. 
All the diffusion-based methods, MDM, MotionDiffuse, MLD, StableMoFusion, and the VAE-based TMR are trained well and converge easily, following the setting in Sec.~\ref{subsec: baseline}.
However, when it comes to the VQVAE-based methods, models perform badly in the first stage of training VQVAE, and the reconstruction results on our test set are far from good as shown in Fig.~\ref{fig:vqvae}. Thus we evaluate only the other methods on our datasets.

\begin{table*}[t!]
 \caption{Quantitative results on the RoleMotion test set. The right arrow  $\rightarrow$ means the closer to real motion the better. \textcolor{red}{Red} and \textcolor{blue}{Blue} indicate the best and the second best result. Models are trained with {\bf body} data, and evaluated on {\bf body} track. 
 R-Precision is calculated in batchsize of 256.
 All evaluations are conducted with 10 runs.}
 \vspace{-2mm}
\label{tab:benchmark-MB-EB}
  \centering
  
\resizebox{\linewidth}{!}{
  \begin{tabular}{lcccccc}
    \toprule
    \multirow{2}{*}{\centering Method} & 
    \multirow{2}{*}{\centering FID $\downarrow$} & 
    \multicolumn{3}{c}{R-Precision$\uparrow$ }& 
    \multirow{2}{*}{\centering Diversity $\rightarrow$}  & 
    \multirow{2}{*}{\centering Multi-modality $\uparrow$} 
    \\
    \cline{3-5}
    && top1 & top2 & top3 & & \\
    \midrule
    Real & -  & $ 0.802^{\pm{ .005}}$ &$ 0.902^{\pm{ .002}}$ &$ 0.931^{\pm{ .001}}$  & $19.121^{\pm{ .058}} $& - \\
    \midrule

    MDM(DDPM1000)~\cite{tevet2023human} & $ 1.871^{\pm{ .016}}$ & $ 0.809^{\pm{ .006}}$ &$ 0.905^{\pm{ .002}}$ &$ 0.938^{\pm{ .003}}$  &$ 19.200^{\pm{ .035}}$& $ 0.201^{\pm{ .024}}$  \\
    
    MLD(DDIM50)~\cite{chen2023executing} & $5.244^{\pm{.118}}$ & ${0.709}^{ \pm .003}$ & ${0.833}^{ \pm .006}$ & $0.875^{\pm{.007}}$ & $19.066^{\pm{.021}}$ & {1.471}$^{\pm{.120}}$  \\ 
 
    MotionDiffuse(DDPM1000)~\cite{zhang2022motiondiffuse} &  $ 1.859^{\pm{ .015}}$ &$ 0.852^{\pm{ .004}}$ &$ 0.935^{\pm{ .002}}$ &$ 0.957^{\pm{ .002}}$ & $ 19.160^{\pm{ .038}}$ & $ 0.177^{\pm{ .022}}$  \\

    TMR~\cite{petrovich23tmr}  & $\textcolor{blue}{1.772}^{\pm{.006}}$ & $\textcolor{red}{0.865}^{\pm{.004}}$ & $ \textcolor{red}{0.948}^{\pm{.001}}$ &$\textcolor{red}{0.970}^{\pm{.002}}$ & ${19.193}^{\pm{.051}}$ & 
    $0.063^{\pm{.025}}$  \\
    MMM~\cite{pinyoanuntapong2024mmm} &  $ 7.558^{\pm{ .041}}$ & $ 0.573^{\pm{ .005}}$ &$ 0.682^{\pm{ .005}}$ &$ 0.734^{\pm{ .004}}$  & $ 19.100^{\pm{ .050}}$ &   \\

    BAD~\cite{hosseyni2025bad} &  $ 7.430^{\pm{ .092}}$  & $ 0.575^{\pm{ .004}}$ &$ 0.682^{\pm{ .005}}$ &$ 0.731^{\pm{ .004}}$  & $ 19.110^{\pm{ .048}}$ &   \\
    
    StableMoFusion(DPMSolver50)~\cite{huang2024stablemofusion} &$ 2.140^{\pm{ .011}}$ &  $ 0.843^{\pm{ .003}}$ &$ 0.933^{\pm{ .003}}$ &$ 0.959^{\pm{ .002}}$   & $ 19.175^{\pm{ .049}}$ & $ 0.202^{\pm{ .022}}$\\
   
    StableMoFusion(DDPM1000)~\cite{huang2024stablemofusion} &$ \textcolor{red}{1.709}^{\pm{ .007}}$ &  $ \textcolor{blue}{0.846}^{\pm{ .003}}$ &$\textcolor{blue} {0.938}^{\pm{ .002}}$ &$ \textcolor{blue}{0.962}^{\pm{ .001}}$& $ 19.195^{\pm{ .047}}$ & $ 0.197^{\pm{ .022}}$\\

    \bottomrule
  \end{tabular}
 }
\end{table*}

\noindent \textbf{Main results.}
We first evaluate the models trained solely with body data on body motion generation. As shown in Tab.~\ref{tab:benchmark-MB-EB}, the FID and R-Precision of MLD in RoleMotion decline severely compared to HumanML3D. Other methods are relatively stable, and StableMoFusion achieves the best FID. Noticeably, RoleMotion exhibits the strongest diversity({\bf 19.121}) than other datasets like Motion-X({\bf 17.683}) and HumanML3D({\bf 15.050}) in the row of real(GT)\footnote{All the diversities are calculated by our evaluator.}.

Second, we evaluate the models trained with whole-body data, including hands, on generating both body and hand motion. As listed in Tab.~\ref{tab:benchmark-MBH-EBH}, the performance of TMR has greatly deteriorated. StableMoFusion keeps to be stable that the version of DDPM1000 and the two-stage version rank first and second in terms of FID and R-Precision.

\noindent \textbf{Influence of training body and hands together.}
To explore the influence of training with body and hands together on body motion generation, we train models on concatenated body and hand data and evaluate only body track in generated output. The results in Tab.~\ref{tab:benchmark-MBH-EB} reveal some interesting insights. Compared with models trained only with body data, it could be found that performance is universally impacted. While diffusion-based methods generally maintain the ranking, the performance of TMR deteriorates even harder than in Tab.~\ref{tab:benchmark-MBH-EBH}. 
Meanwhile, the model trained using two-stage schema achieves the finest results.
We infer that the numerous joints of hands squeeze the learning space of body joints, models like TMR trained with contrastive loss are greatly influenced compared to diffusion-based models.
Another insight is that results evaluated on both body and hand seem to be more decent, and this implies that the numerous joints of hands are also suppressing body during evaluation. The result that two-stage model has weaker FID and R-Precision than the one-stage model also verifies this.
Thus it is unreasonable to evaluate the body\&hand motion generation with a single evaluator trained with body\&hand. It is essential to train an evaluator of body and evaluate the body, and train an evaluator of body\&hand and evaluate body\&hand as an indicator of hand motion.

\begin{table*}[t!]
 \caption{Quantitative results of models trained with {\bf body} and {\bf hand} data, and only evaluated on {\bf body} track.}
 \vspace{-3mm}
\label{tab:benchmark-MBH-EB}
  \centering
  
\resizebox{\linewidth}{!}{
  \begin{tabular}{lcccccc}
    \toprule
    \multirow{2}{*}{\centering Method} & 
    \multirow{2}{*}{\centering FID $\downarrow$} & 
    \multicolumn{3}{c}{R-Precision$\uparrow$ }& 
    \multirow{2}{*}{\centering Diversity $\rightarrow$}  & 
    \multirow{2}{*}{\centering Multi-modality $\uparrow$} 
    \\
    \cline{3-5}
    & & top1 & top2 & top3 & & \\
    \midrule
     Real & -  & $ 0.802^{\pm{ .005}}$ &$ 0.902^{\pm{ .002}}$ &$ 0.931^{\pm{ .001}}$  & $19.121^{\pm{ .058}} $& - \\
    \midrule

    MDM(DDPM1000)~\cite{tevet2023human} & $ 2.236^{\pm{ .022}}$& $ 0.781^{\pm{ .003}}$ &$ 0.884^{\pm{ .004}}$ &$ 0.917^{\pm{ .002}}$ & $ 19.151^{\pm{ .049}}$ & $ 0.330^{\pm{ .041}}$  \\
    
    MLD(DDIM50)~\cite{chen2023executing} & $9.016^{\pm{.117}}$ & ${0.595}^{ \pm .004}$ & ${0.722}^{ \pm .007}$ & $0.770^{\pm{.007}}$ & $19.011^{\pm{.043}}$ & {2.129}$^{\pm{.138}}$  \\ 
 
    MotionDiffuse(DDPM1000)~\cite{zhang2022motiondiffuse} & $ 2.985^{\pm{ .017}}$ & $ 0.807^{\pm{ .002}}$ &$ 0.905^{\pm{ .003}}$ &$ 0.936^{\pm{ .001}}$  & $ 19.129^{\pm{ .051}}$  & $ 0.144^{\pm{ .019}}$  \\

    TMR~\cite{petrovich23tmr}  & $ 2.497^{\pm{ .009}}$ &$ 0.838^{\pm{ .003}}$ &$ 0.932^{\pm{ .002}}$ &$ 0.958^{\pm{ .001}}$ & $ 19.148^{\pm{ .049}}$ & 
    $1.450^{\pm{.058}}$  \\
    
    
    

    MMM~\cite{pinyoanuntapong2024mmm} &   $ 7.951^{\pm{ .070}}$ & $ 0.392^{\pm{ .003}}$ &$ 0.480^{\pm{ .002}}$ &$ 0.539^{\pm{ .003}}$  & $ 19.087^{\pm{ .057}}$ &    \\

    BAD~\cite{hosseyni2025bad} &   $ 8.474^{\pm{ .073}}$ & $ 0.535^{\pm{ .003}}$ &$ 0.631^{\pm{ .004}}$ &$ 0.677^{\pm{ .005}}$  & $ 19.046^{\pm{ .063}}$ &    \\
    
    StableMoFusion(DPMSolver50)~\cite{huang2024stablemofusion} &$ 2.454^{\pm{ .014}}$& $ 0.813^{\pm{ .005}}$ &$ 0.920^{\pm{ .002}}$ &$ 0.949^{\pm{ .002}}$ & $ 19.072^{\pm{ .040}}$ & $ 0.190^{\pm{ .025}}$\\
    
    StableMoFusion(DDPM1000)~\cite{huang2024stablemofusion} & $ \textcolor{blue}{2.042}^{\pm{ .012}}$ &  $ \textcolor{blue}{0.820}^{\pm{ .004}}$ &$ \textcolor{blue}{0.928}^{\pm{ .002}}$ &$ \textcolor{blue}{0.956}^{\pm{ .002}}$& $ 19.097^{\pm{ .041}}$ & $ 0.188^{\pm{ .024}}$\\
    \midrule
    Two-stage(DDPM1000) & $ \textcolor{red}{1.719}^{\pm{ .007}}$ &  $ \textcolor{red}{0.849}^{\pm{ .004}}$ &$\textcolor{red} {0.938}^{\pm{ .003}}$ &$ \textcolor{red}{0.962}^{\pm{ .001}}$& $ 19.133^{\pm{ .036}}$ & $ 0.197^{\pm{ .022}}$ \\

    \bottomrule
  \end{tabular}
}
 
\end{table*}

\begin{table*}[t!]
 \caption{Quantitative results of models trained with {\bf body} and {\bf hand} data, and evaluated on {\bf body} and {\bf hand} track.}
 \vspace{-3mm}
\label{tab:benchmark-MBH-EBH}
  \centering
  
\resizebox{\linewidth}{!}{
  \begin{tabular}{lcccccc}
    \toprule
    \multirow{2}{*}{\centering Method} & 
    \multirow{2}{*}{\centering FID $\downarrow$} & 
    \multicolumn{3}{c}{R-Precision$\uparrow$ }& 
    \multirow{2}{*}{\centering Diversity $\rightarrow$}  & 
    \multirow{2}{*}{\centering Multi-modality $\uparrow$} 
    \\
    \cline{3-5}
    & & top1 & top2 & top3 & & \\
    \midrule
     Real & - &$0.828^{\pm{ .004}}$ &$ 0.926^{\pm{ .002}}$ &$ 0.950^{\pm{ .001}}$&
 $ 19.157^{\pm{ .035}}$   &  - \\
    \midrule

    MDM(DDPM1000)~\cite{tevet2023human} & $ 1.786^{\pm{ .023}}$ & $ 0.805^{\pm{ .004}}$ &$ 0.904^{\pm{ .002}}$ &$ 0.934^{\pm{ .002}}$ & $ 19.134^{\pm{ .051}}$ &  $ 0.498^{\pm{ .051}}$  \\
   
    MLD(DDIM50)~\cite{chen2023executing} & $4.941^{\pm{.145}}$ & ${0.690}^{ \pm .009}$ & ${0.807}^{ \pm .007}$ & $0.846^{\pm{.006}}$ & $19.081^{\pm{.045}}$ & {3.990}$^{\pm{.210}}$  \\ 
 
    MotionDiffuse(DDPM1000)~\cite{zhang2022motiondiffuse} & $ 1.760^{\pm{ .012}}$&$ 0.849^{\pm{ .003}}$ &$ 0.935^{\pm{.002}}$ &$ 0.960^{\pm{ .002}}$  & $ 19.151^{\pm{ .051}}$ & $0.269^{\pm{.042}}$   \\

    TMR~\cite{petrovich23tmr}  &  $ 1.762^{\pm{ .013}}$ & $ 0.848^{\pm{ .003}}$ &$ 0.935^{\pm{ .002}}$ &$ 0.960^{\pm{ .001}}$  &$ 19.159^{\pm{ .057}}$  & 
    $ 0.267^{\pm{.030}}$   \\

    MMM~\cite{pinyoanuntapong2024mmm} &   $ 6.533^{\pm{ .095}}$ & $ 0.561^{\pm{ .007}}$ &$ 0.663^{\pm{ .004}}$ &$ 0.713^{\pm{ .004}}$  & $ 19.152^{\pm{ .027}}$ &    \\

    BAD~\cite{hosseyni2025bad} &   $ 6.524^{\pm{ .062}}$ & $ 0.560^{\pm{ .004}}$ &$ 0.664^{\pm{ .004}}$ &$ 0.713^{\pm{ .004}}$  & $ 19.073^{\pm{ .036}}$ &    \\
    
    
    
    
    StableMoFusion(DPMSolver50)~\cite{huang2024stablemofusion} & $ 1.690^{\pm{ .009}}$ & $ 0.864^{\pm{ .005}}$ &$ {0.952}^{\pm{ .002}}$ &$ {0.973}^{\pm{ .001}}$ & $ 19.122^{\pm{ .053}}$ & $ 0.316^{\pm{ .032}}$\\
    
    StableMoFusion(DDPM1000)~\cite{huang2024stablemofusion} &$ \textcolor{blue}{1.532}^{\pm{ .008}}$&$\textcolor{blue}{0.867}^{\pm{ .004}}$ &$\textcolor{blue}{ 0.952}^{\pm{ .003}}$ &$ \textcolor{red}{0.973}^{\pm{ .002}}$  & $  19.133^{\pm{ .050}}$ &  $ 0.311^{\pm{ .031}}$\\
    \midrule
    Two-stage(DDPM1000) &  $ \textcolor{red}{1.426}^{\pm{ .006}}$ & $\textcolor{red}{0.872}^{\pm{ .003}}$ &$ \textcolor{red}{0.953}^{\pm{ .002}}$ &$\textcolor{blue}{ 0.970}^{\pm{ .001}}$ & $ 19.168^{\pm{ .037}}$ & $ 0.747^{\pm{ .046}}$ \\

    \bottomrule
  \end{tabular}
}
 
\end{table*}

\subsection{Visualizations}
\noindent\textbf{Fine-grained language driven human motion synthesis.}
To display the quality of data and annotation, we compare models trained on HumanML3D and RoleMotion. 
We randomly create motion descriptions that need to specify the motion details while keeping all the action categories included in both datasets for fairness. 
As illustrated in Fig.~\ref{fig:fine_grained}, almost all the action details specified in the text are accurately conducted by model trained on RoleMotion,  while motions of model trained on HumanML3D tend to be unnatural and contain many errors regarding body states, side of part and direction. This implies that the quality of motion data and granularity of annotation are extremely important for accurate motion generation.




\vspace{-3mm}
\section{Conclusion }
\vspace{-1mm}
This paper presents RoleMotion, a large-scale human motion dataset featuring extensive role-playing motions across diverse scenarios. The data is meticulously designed, collected, and annotated with fine-grained textual descriptions. Experiments show that models trained on RoleMotion generate higher-quality motions with better textual alignment compared to existing datasets. We further establish a reliable evaluation benchmark for text-to-motion tasks. Finally, we explore the interplay between body and hand motion synthesis, advocating separate training and evaluation for each modality.
Nevertheless, convincingly evaluating hand motion remains challenging, as it requires simultaneous alignment with both text  and body movement.
{
    \small
    \bibliographystyle{ieeenat_fullname}
    \bibliography{main}
}


\end{document}